\documentclass[10pt,twocolumn,letterpaper]{article}

\usepackage{cvpr}
\usepackage{times}
\usepackage{epsfig}
\usepackage{graphicx}
\usepackage{amsmath}
\usepackage{amssymb}

\usepackage{subfig}
\usepackage{caption}

\usepackage{booktabs}

\newcommand{\mytilde}{\raise.17ex\hbox{$\scriptstyle\mathtt{\sim}$}}

\usepackage{bm}

\usepackage[pagebackref=true,breaklinks=true,letterpaper=true,colorlinks,bookmarks=false]{hyperref}

\cvprfinalcopy 

\def\cvprPaperID{748}

\ifcvprfinal\fi
\begin{document}

\title{Oriented Edge Forests for Boundary Detection}
\author{Sam Hallman \quad \quad \quad Charless C. Fowlkes\\
Department of Computer Science\\
University of California, Irvine\\
{\tt\small \{shallman,fowlkes\}@ics.uci.edu}
}

\maketitle

\begin{abstract}
We present a simple, efficient model for learning boundary detection based 
on a random forest classifier.  Our approach combines (1) efficient clustering
of training examples based on a simple partitioning of the space of local edge
orientations and (2) scale-dependent calibration of individual tree output
probabilities prior to multiscale combination. The resulting model outperforms
published results on the challenging BSDS500 boundary detection benchmark.
Further, on large datasets our model requires substantially less memory for
training and speeds up training time by a factor of 10 over the structured
forest model.
\footnote{This work was supported by NSF DBI-1053036, DBI-1262547, and IIS-1253538}
\end{abstract}

\section{Introduction}

Accurately detecting boundaries between objects and other regions in images has
been a long standing goal since the early days of computer vision.  Accurate
boundary estimation is an important first step for segmentation and detection
of objects in a scene and boundaries provide useful information about the shape
and identity of those objects.  Early work such as the Canny edge
detector~\cite{canny} focused on detecting brightness edges, estimating their
orientation~\cite{billfreemansteerablefilters} and analyzing the theoretical limits of
detection in the presence of image noise.  However, simple brightness or color
gradients are insufficient for handling many natural scenes where local
gradients are dominated by fine scale clutter and texture arising from surface
roughness and varying albedo. 

Modern boundary detectors, such as~\cite{Pb}, have emphasized the importance of
suppressing such responses by explicit oriented analysis of higher order
statistics which are robust to such local variation. These statistics can be
captured in a variety of ways, \eg via textons~\cite{LeungMalik}, sparse
coding~\cite{SCG}, or measures of self-similarity~\cite{Leordeanu}. Such
boundary detectors also generally benefit from global normalization provided by
graph-spectral analysis~\cite{gPbPAMI} or ultra-metric consistency~\cite{UCM} which
enforce closure, boosting the contrast of contours that completely enclose
salient regions.

Recently, focus has turned to methods that learn appropriate feature
representations from training data rather than relying on carefully
hand-designed texture and brightness contrast measures.  For example,
\cite{SCG} learns weightings for each sparse code channel and hypothesized edge
orientation while \cite{BEL,SketchTokens} predict the probability of a boundary
at an image location using a cascade or randomized decision forest built over
simple image features. Taking this one step further, the work of
\cite{MaireYuPeronaACCV2014} and \cite{PiotrICCV} learn not only input features
but also the output space using sparse coding or structured-output decision
forests respectively.  While these approaches haven't yielded huge gains in
boundary detection accuracy, they are appealing in that they can adapt to other
domains (\eg, learning input features for boundary detection in RGB-D images
\cite{SCG,PiotrICCV} or predicting semantic segmentation
outputs~\cite{MaireYuPeronaACCV2014}).  On the other hand, a key difficulty
with these highly non-parametric approaches is that it is difficult to control
what is going on ``under the hood'' and to understand why they fail or succeed
where they do.  Like a fancy new car, they are great when they work, but if ever
stranded on a remote roadside, one suddenly discovers there are very few user
serviceable parts inside.

\begin{figure}
  \centering
  \includegraphics[width=\linewidth]{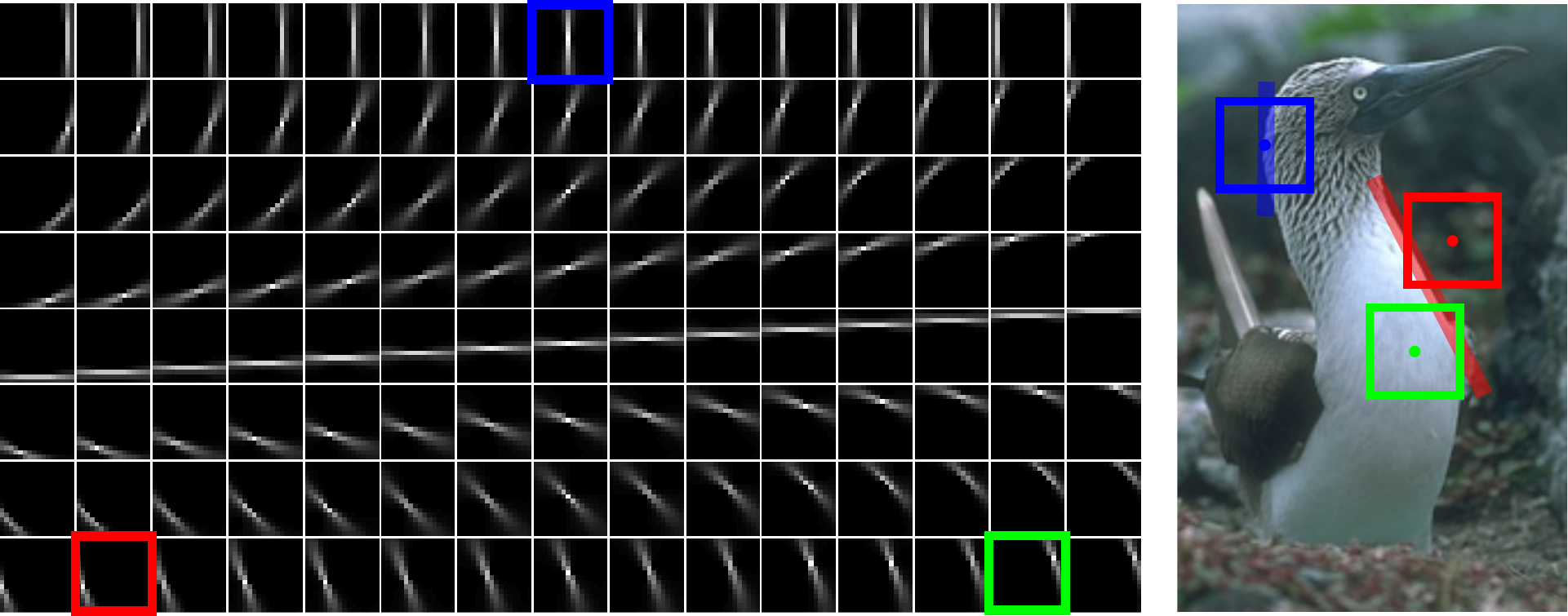}
  \caption{Our boundary detector consists of a decision forest that analyzes
  local patches and outputs probability distributions over the space of 
  oriented edges passing through the patch.  This space is indexed by 
  orientation and signed distance to the edge $(d,\theta)$.  These local
  predictions are calibrated and fused over an image pyramid to yield
  a final oriented boundary map. 
  \label{fig:angrybird}
  }

\vspace{-0.165cm}
\end{figure}

In this paper we to take a step back from non-parametric outputs and instead
apply the robust machinery of randomized decision forests to the simple task of
accurately detecting straight-line boundaries at different candidate
orientations and positions within a small image patch. Although this ignores a
large number of interesting possibilities such as curved edges and junctions,
it should certainly suffice for most small patches of images containing big,
smooth objects. We show that such a model, appropriately calibrated and
averaged across a small number of scales, along with local sharpening of edge
predictions outperforms the best reported results on the BSDS500 boundary
detection benchmark.

The rest of the paper is structured as follows. In Section~\ref{sec:clust}, we
describe our method for partitioning the space of possible oriented edge
patterns within a patch.  This leads to a simple, discrete labeling over local
edge structures. In Section~\ref{sec:forest}, we discuss how to use this
discrete labeling to train a random forest to predict edge structure within a
patch, and describe a calibration procedure for improving the posterior
distributions emitted by the forest. Section~\ref{sec:synthesis} then describes
how to map the distributions computed over the image into a final, high-quality
edge map. Finally, in Section~\ref{sec:experiments} we show experimental results
on the BSDS500 boundary detection benchmark.

\section{Clustering Edges}
\label{sec:clust}

From a ground truth boundary image, we categorize a $p \times p$ patch either
as containing no boundary (background) or as belonging to one of a fixed number
of edge categories. A patch is considered background if its center is more than
$p/2$ pixels away from an edge, in which case the patch contains little to no
edge pixels.

Non-background patches are distinguished according to the distance $d$ and
orientation $\theta$ of the edge pixel closest to the patch center.  Thus,
patches with $d=0$ have an edge running through the center, and by definition
$d$ is never greater than $p/2$.  We choose a canonical orientation for each
edge so that $\theta$ lies in the interval $(-\pi/2,\pi/2]$.  To distinguish
between patches on different sides of an edge with the same orientation, we
utilized signed distances $d \in (-p/2,p/2)$.  This yields a parameter pair
$(d,\theta)$ for each non-background patch.

Figure~\ref{fig:angrybird} shows this two dimensional space of patches.  It is
worth noting that this space can be given an interesting topology.  Since
orientation is periodic, a straight edge with parameter $(d,\theta)$ appears
identical to one with parameter $(-d,\theta+\pi)$. One can thus identify the
top and bottom edges of the space in Figure 1, introducing a half-twist to
yield a M{\"o}bius strip whose boundary is $\{(d,\theta) : |d|=p/2\}$.%
\footnote{One could also parameterize lines by angle $\theta \in (-\pi,\pi]$ and
unsigned distance $d \geq 0$.  However, this space has a singularity at $d=0$
where patches $(0,\theta)$ and $(0,\theta+\pi)$ are indistinguishable to an
edge detector but have different angle parameters.  Our
parameterization is convenient since it assigns unique coordinates to each
line and is smooth everywhere.}

From a ground-truth edge map, computing the distance between a patch center
and the nearest edge pixel $q$ is straightforward. To be useful, the estimate
of $\theta$ should reflect the dominant edge direction around $q$, and be
robust to small directional changes at $q$. To accomplish this, we first link
all edge pixels in a ground-truth boundary map into edge lists, breaking lists
into sublists where junctions occur. We then measure the angle at $q$ by
fitting a polynomial to the points around $q$ that are in the same list. In our
experiments we use a fitting window of $\pm6$ pixels.

Because annotators sometimes attempt to trace out extremely fine detail around
an object, boundary annotations will occasionally include very short, isolated
``spur'' edges protruding from longer contours. Where these occur, estimates of
$\theta$ can suffer. We remove all such edges provided that they are shorter
than $7$ pixels in length. Using standard morphological operations we also fill
holes if they exist and thin the result to ensure that all lines are a single
pixel thick.

\paragraph{Collecting training data}
We binned the space of distances $d$ and angles $\theta$ into $n$ and $m$ bins,
respectively. Thus every non-background patch was assigned to a discrete
label $k$ out of $K=nm$ possible labels. This discrete label space allows for
easy application of a variety of off-the-shelf supervised learning algorithms.  

In our experiments we used a patch size of $16\times16$ pixels, so that
distances satisfy $|d|<p/2=8$. It is natural to set the distance bins one pixel
apart, so that $d$ falls into one of $n=15$ bins. Assigning angles $\theta$ to
one of $m=8$ bins, leaves $K=120$ edge classes plus background.  We chose
the orientation binning so that bins 1 and 5 are centered at 90 and 0 degrees
respectively, as these orientations are especially common in natural images
\cite{imgstats}.  Figure~\ref{fig:angrybird}(a) shows the average ground-truth
edge map for all image patches assigned to each of these clusters.

In our experiments we sampled patches uniformly over image locations and over
labelings derived from multiple ground-truth segmentations of that image.
Since our approach ultimately predicts a $(d,\theta)$ parameter for each
non-background image patch, it does not explicitly model patches containing
junctions or thin structures involving more than two segments.  In practice,
such events are relatively rare.  In the BSDS500 training dataset, patches
containing more than two segments constitute less than 8\% of image patches and
only 27\% of all non-background patches. To simplify the learning problem
faced by the local classifier, we only utilize patches that contain one or two
segments for training.

\section{Oriented Edge Forest}
\label{sec:forest}
Using the labeling procedure outlined in Section~\ref{sec:clust}, we can build
a training dataset comprised of color image patches $\mathbf{x}$, each with a
corresponding edge cluster assignment $\mathbf{y} \in \{0,1,\ldots,K\}$ where
$K$ is the number of edge clusters and $\mathbf{y}=0$ represents the background
or ``no boundary'' class.  Inspired by the recent success of random decision
forests for edge detection \cite{SketchTokens,PiotrICCV}, we train a random
forest classifier to learn a mapping from patches to this label set.
In this section we discuss forest training and calibration procedures that
yield high-quality edge probability estimates.

\begin{figure}[t]
\begin{center}
   \includegraphics[width=\linewidth]{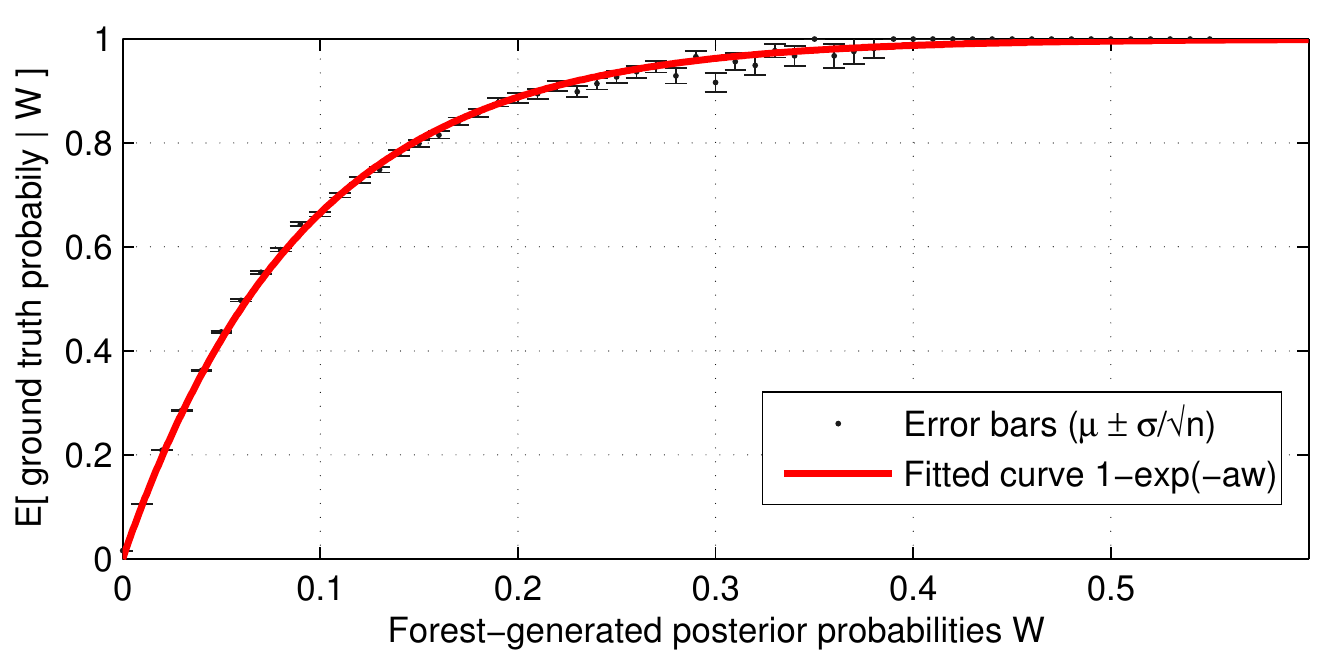}
\end{center}
\caption{Reliability plot showing the empirical probability of a ground-truth
edge label as a function of the score output by the forest computed over a set 
of validation images.  Error bars show standard error in the empirical
expectation.  Red curve shows a simple functional fit $1-\exp(-\beta w)$
which appears to match the empirical distribution well.  We use this
estimated function (one scalar parameter $\beta$ per scale) to calibrate the
distribution of scores over different edges $(d,\theta)$ predicted by the
forest.  Performing this calibration prior to combining and compositing
predictions across scales improves final performance.  }
\label{fig:calib}
\end{figure}

\paragraph{Randomized Decision Forests}
Random forests are a popular ensemble method in which randomized decision trees
are combined to produce a strong classifier.  Trees are made random through
bagging and/or randomized node optimization \cite{criminisi2013decision}, in
which the binary splits at the nodes of the tree are limited to using only a
random subset of features.

In our framework, the output label space predicted by the forest is a small
discrete set ($K$ possible edge orientations and locations relative to the
center of the patch or background) and may be treated simply as a k-way
classification problem.  When training a given decision tree, features are
selected and split thresholds are chosen to optimize the Gini impurity
measure~\cite{breiman1984classification}.  In practice we find that the
particular choice of class purity metric does not have a noticeable impact on
performance.  We did find it important to have balanced training data across
classes and used an equal number training examples per class.

\paragraph{Image Features} We adopt the same feature extraction process used
in \cite{PiotrICCV}. In this approach, images are transformed into a set of
feature channels, and the descriptor for a patch is computed simply by cropping
from the corresponding window in the array of feature channels. These features
are comprised of color and gradient channels, and are downsampled by a factor
of 2. Binary splits performed at the tree nodes are accomplished by
thresholding either a pixel read from a channel or the difference between two
pixels from the same channel. See \cite{PiotrICCV} for full details.

\paragraph{Ensemble Averaging}
Equipped with a forest trained to recognize oriented edge patterns, the next
step is to apply the forest over the input image. We have found that the
details of how we fuse the predictions of different trees can have a
significant effect on performance.  Two standard approaches to combining 
the output of a ensemble of classifiers are averaging and voting.

For a given test image patch $\mathbf{x}$, each individual tree $t$ produces an
estimate $p_t(k|\mathbf{x})$ of the posterior distribution over the $K+1$ class
labels based on the empirical distribution observed during training. We would
like to combine these individual estimates into a final predicted score vector
$\mathbf{w}(k|\mathbf{x})$. The most obvious way to combine the tree outputs is
\emph{averaging}
\begin{equation}
  \textbf{w}(k|\mathbf{x}) =
    \frac{1}{T} \sum_{t=1}^T p_t(k|\mathbf{x}), \qquad k=1,...,K
  \label{eqn:avg}
\end{equation}
An alternative, often used for ensembles of classifiers which only output
class labels instead of posteriors is \emph{voting}
\begin{equation}
  \textbf{w}(k|\mathbf{x}) =
    \frac{1}{T} \sum_{t=1}^T \mathbf{1}_{[k=\arg\max_k p_t(k|\mathbf{x})]}
  \label{eqn:vote}
\end{equation}
where $\mathbf{1}$ is the indicator function.

In general, we find that averaging provides somewhat better detection accuracy
than voting, presumably because the votes carry less information than the full
posterior distribution (see Section~\ref{sec:experiments}).  One disadvantage
of averaging is that it requires one to maintain in memory all of the empirical
distributions $p$ at every leaf of every tree. Voting not only requires less
storage for the forest but also reduces runtime.  Constructing $\mathbf{w}$
via averaging requires $O(KT)$ while voting only requires $O(T)$.  The
resulting $\mathbf{w}$ is also sparse which can lead to substantial speed
improvements in the edge fusion steps described below
(Section~\ref{sec:synthesis}).  Voting may thus be an efficient alternative for
time-critical applications.

\paragraph{Calibration}

In order to fuse edge predictions across different scales within an image and
provide boundary maps whose values can be meaningfully compared between images,
we would like the scores $\mathbf{w}$ to be accurately calibrated. Ideally the
scores $\mathbf{w}$ output for a given patch would be the true posterior
probability over edge types for that patch. Let $\mathbf{x}$ be a patch sampled
from the dataset and $y$ the true edge label for that patch. If the scores
$\mathbf{w}(k|\mathbf{x})$ output by the classifier are calibrated then we would
expect that
\begin{equation}
  P(y = k \, | \, \mathbf{w}(k|\mathbf{x}) = s) = s
\end{equation}

To evaluate calibration, we extracted a ground-truth label indicator vector for
every labeled image patch in a held-out set of validation patches
$\{(\mathbf{x}_i,y_i)\}$.\footnote{When multiple humans segmentations generated
conflicting labels for the patch, we averaged them to produce a ``soft''
non-binary label vector.} We then computed the empirical expectation of how
often a particular label $k$ was correct for those patches that received a
particular score $s$.
\begin{align}
  P(y =k \,|\, \mathbf{w}(k|\mathbf{x}) = s) \approx
    \frac{1}{|B(k,s)|} \sum_{i \in B(k,s)} \mathbf{1}_{[y_i=k]} 
\end{align}
where
\[
B(k,s) = \{i : \mathbf{w}(k|\mathbf{x}) \in [s\pm \epsilon]\} \\
\]
is a bin of width $2\epsilon$ centered at $s$.

Figure \ref{fig:calib} shows the resulting reliability plot, aggregated over
non-background patches.  Results were very similar for individual edge labels.
While one might expect that a forest trained to minimize entropy of the 
posterior predictions would tend to be overconfident, we found that the forest
average scores for non-background patches actually tended to underestimate the
true posterior! This remained true regardless of whether we used voting or
averaging.

Previous work has used logistic regression in order to calibrate classifier
output scores \cite{plattscaling}.  For
the oriented edge forest, we found that this miscalibration for non-background
labels is much better fit by an exponential
\begin{equation}
  \hat{\textbf{w}}(k|\mathbf{x}) =
    f_\beta(\textbf{w}(k|\mathbf{x})) =
      1 - \exp(-\beta \textbf{w}(k|\mathbf{x}))
\end{equation}
where $\beta$ is a scalar.  We fitted this function directly to the binary
indicator vectors $\mathbf{1}_{[y_i=k]}$ rather than binned averages in order
to give equal weight to each training example.

We also explored a wide variety of other calibration models including
sigmoid-shaped functions such as $\tanh$, richer models that 
fit an independent parameter $\beta_k$ per class label, and 
joint calibration across all class labels. We even considered a non-parametric
approach in which we treated the 120-D ground truth label vectors as structured
labels and trained an additional structured random forest \cite{PiotrPAMI}.
We found that using a single scalar $\beta$ for all non-background scores is
highly efficient\footnote{For a sparse voting implementation, one can do
nearly as well using the fast approximation $f(\textbf{w}) =
\min\{1,\textbf{w}\}$.} and performed as well as any calibration scheme we
tried.  When performing multiscale fusion (Section~\ref{sec:synthesis}), we fit
a distinct $\beta$ for each scale, the values of which typically ranged from 6
to 10.

\begin{figure}[t]
  \captionsetup[subfigure]{labelformat=empty}
  \centering
  \subfloat[patch]{%
    \begin{tabular}{c}
      \includegraphics[width=0.15\linewidth]{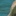} \\
      \includegraphics[width=0.15\linewidth]{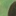} \\
      \includegraphics[width=0.15\linewidth]{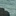}
    \end{tabular}
  }%
  \subfloat[$\mathrm{sh}=0$]{%
    \begin{tabular}{c}
      \includegraphics[width=0.15\linewidth]{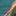} \\
      \includegraphics[width=0.15\linewidth]{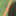} \\
      \includegraphics[width=0.15\linewidth]{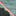}
    \end{tabular}
  }%
  \subfloat[$\mathrm{sh}=1$]{%
    \begin{tabular}{c}
      \includegraphics[width=0.15\linewidth]{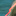} \\
      \includegraphics[width=0.15\linewidth]{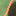} \\
      \includegraphics[width=0.15\linewidth]{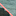}
    \end{tabular}
  }%
  \subfloat[$\mathrm{sh}=2$]{%
    \begin{tabular}{c}
      \includegraphics[width=0.15\linewidth]{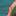} \\
      \includegraphics[width=0.15\linewidth]{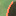} \\
      \includegraphics[width=0.15\linewidth]{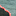}
    \end{tabular}
  }%
  \caption{Examples of sharpening a single predicted edge patch label based
  on underlying image evidence.  The patch is resegmented based on the 
  initial straight edge label (2nd column) by reassigning pixels near the 
  boundary to the region with more similar mean RGB value.
  }
  \label{fig:sharpeningpatches}
\end{figure}

\section{Edge Fusion}
\label{sec:synthesis}

Having applied the forest over the input image, we are left with a collection
of calibrated probability estimates $\hat{\mathbf{w}}$ at every spatial
position.  Because these distributions express the likelihood of both centered
($d=0$) as well as distant, off-center ($d\neq0$) edges, the probability of
boundary at a given location is necessarily determined by the tree predictions
over an entire neighborhood around that location. In this section, we describe
how to resolve these probabilities into a single, coherent image of boundary
strengths. The end result will be an oriented signal $E(x,y,\theta)$ that
specifies the probability of boundary at location $(x,y)$ in the binned
direction $\theta$.

\paragraph{Edge sharpening} By focusing on oriented lines, our detector is
trained to recognize coarse edge statistics but cannot predict more detailed
structure, \eg local curvature or wiggles of a few pixels in a contour.
As the size of the analyzed patch increases relative to the size of an object,
the straight line assumption becomes a less accurate representation of the
shape.  In order to provide a more detailed prediction of the contour shape, we
utilize a local segmentation procedure similar to the sharpening method
introduced by Doll\'{a}r and Zitnick~\cite{PiotrPAMI}. This is similar in
spirit to the notion of ``Edge Focusing'' \cite{FredrikBergholmTPAMI1987} in
which coarse-to-fine tracking utilizes edge contrast measured at a coarse scale
but contour shape derived from fine scale measurements.

Consider a hypothesized (straight) edge predicted by the forest at a given
location.  We compute the mean RGB color of the pixels on each side of the
hypothesized edge inside a $16\times16$ pixel patch centered at the location.
We then re-segment pixels inside the patch by assigning them to one of these
two cluster means. To prevent the local segmentation from differing wildly with
the original oriented line predicted by the forest, we only reassign pixels
within 1 or 2 pixels distance from the hypothesized segment boundary.
We will use the notation $M_{(x,y,k)}(i,j)$ to denote the sharpened binary edge
mask of type $k = (d,\theta)$ computed for a patch centered at location $(x,y)$
in an input image.  Figure~\ref{fig:sharpeningpatches} shows examples of
individual patches along with the resulting mask $M$ for more and less
aggressive sharpening. 

\paragraph{Compositing}
Given local estimates of the likelihood (calibrated scores $\hat{\mathbf{w}}$)
and precise boundary shapes (sharpened masks $M$) for each image patch, we
predict whether a location $(x,y)$ is on a boundary by averaging over
patch predictions for all patches that include the given location.  Using the
convention that $M_{(x,y,k)}(0,0)$ is the center of a given edge
mask and indexing $\hat{\mathbf{w}}$ by the coordinates of each patch in
the image, we can write this formally as
\begin{equation*}
E(x,y,\theta) =\sum_{k \in \{(d,\theta) \forall d\}} \sum_{(i,j)\in O_{xy}}  \hat{\mathbf{w}}(i,j,k)  M_{(i,j,k)}(x-i,y-j)
\label{eqn:paste}
\end{equation*}
where $O_{xy}$ are the coordinates of patches overlapping $x,y$ and $k$ ranges
over all predicted labels which are compatible with orientation $\theta$.%
\footnote{Note that if we do not perform any sharpening on the edge masks, then
$M$ is the same at every image location $(i,j)$ and the resulting operation is
simply a correlation of $\hat{\mathbf{w}}$ with $M$ summed over channels $k$.}

\begin{figure}[ht]
\begin{center}
   \includegraphics[width=0.7\linewidth]{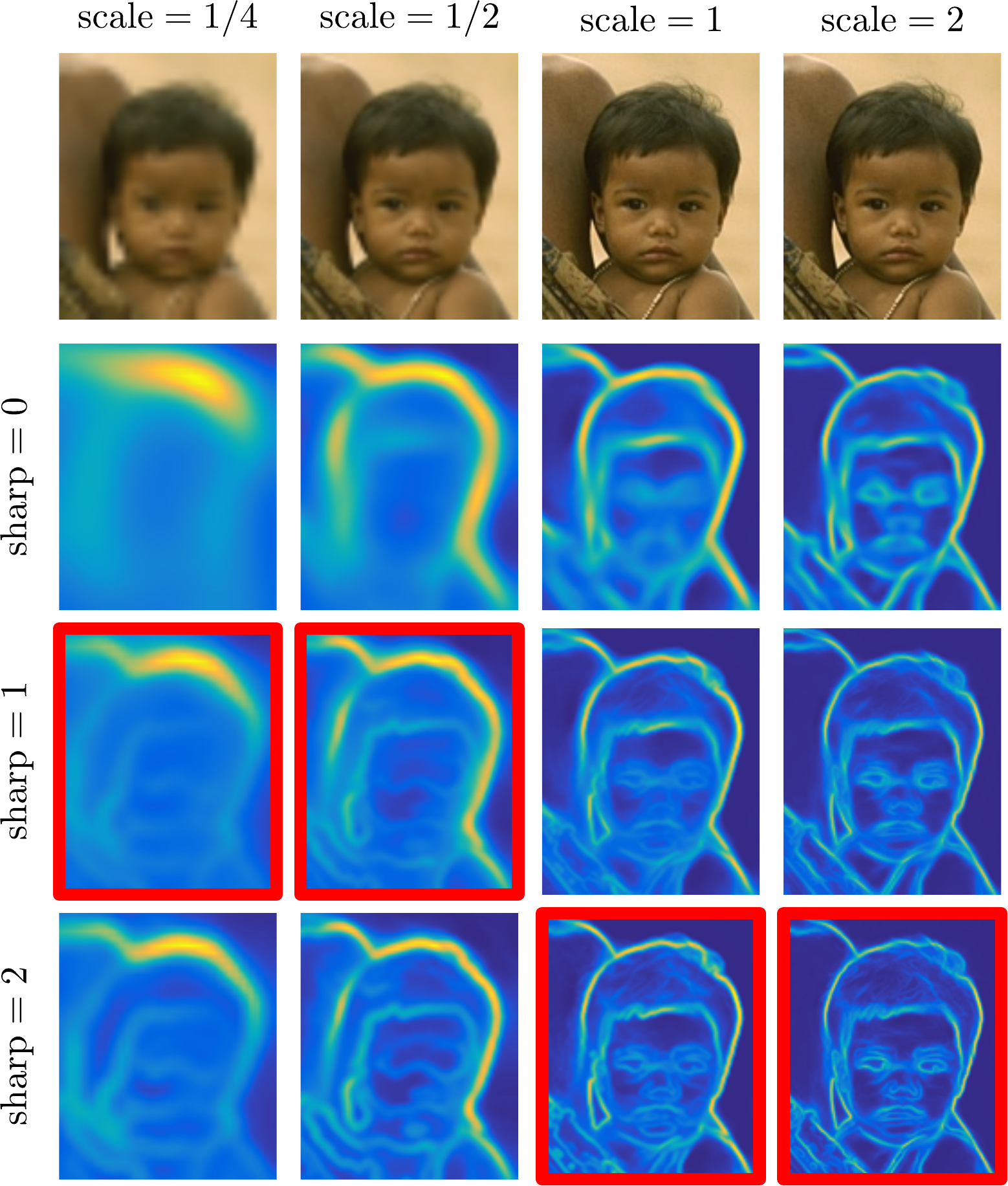}
\end{center}
   \caption{
  The output of the forest when run at different scales (by down/up-sampling
  the input image and with different degrees of edge sharpening).  Running the
  forest on a low-resolution version of the image yields blurry detections that
  respond to coarse scale image structure.  Sharpening allows the spatial
  localization of this broad signal and alignment of predictions made by
  overlapping local classifiers.  We found that coarse scale information is
  best utilized by performing only modest sharpening of the lowest-resolution
  output to allow strong, broad edge responses to combine with finely localized
  edges from higher scales.}
\label{fig:sharp}
\end{figure}

\paragraph{Combining multiple scales} 
The compositing procedure in the previous section can easily be repeated to
produce an $E(x,y,\theta,s)$ for different scaled versions of an input image.
In general, combining results at different scales is known to improve
performance \cite{ren2008multi}.  We apply the detector at four scales. To
detect large-scale edge structure we run at scales $s=1/4,\,1/2$. We find that
at these resolutions heavy sharpening is less desirable (see
Figure~\ref{fig:sharp}). Finer edge structure is discovered at scales $s=1,2$,
and at these scales more aggressive sharpening is preferred. The results are
averaged to produce a final output, as in \cite{PiotrICCV}.  The strengths of
each scale can be seen in benchmark results in Figure~\ref{fig:scalePR}, where
the curves tend toward higher precision and lower recall as $s$ decreases. It
is interesting to note that including $s=2$ is beneficial despite being
dominated everywhere by $s=1$. As lower scales are added, precision increases
but asymptotic recall suffers. Including scale 2 allows us to maintain the
benefits of low scales without the loss in recall.

\section{Experiments}
\label{sec:experiments}


\subsection{Benchmark Performance}
\label{sec:bench}

\begin{figure}[t]
  \begin{center}
    \includegraphics[width=0.9\linewidth]{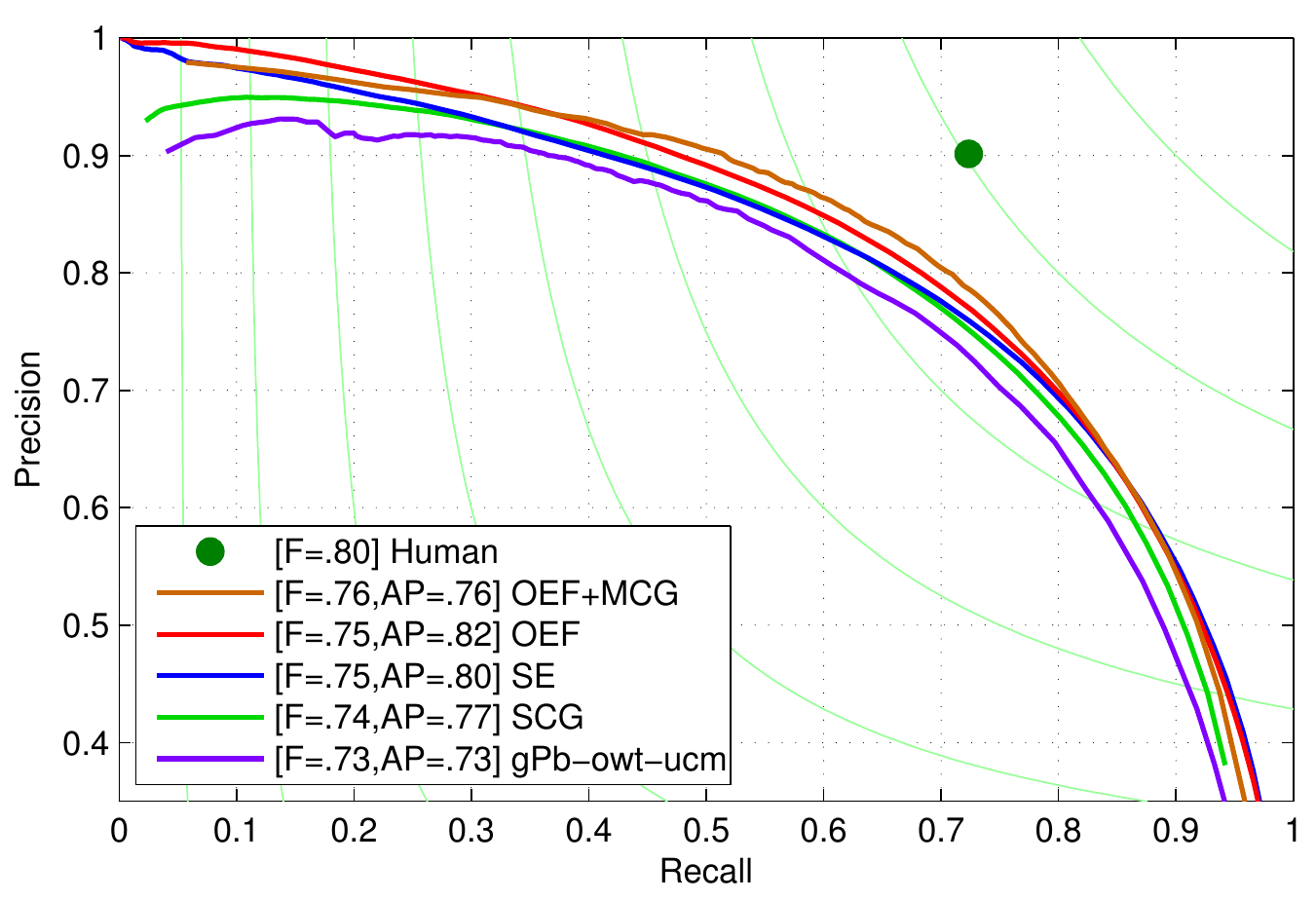}
  \end{center}
  \vspace{-0.1in}
  \caption{Results on BSDS500. Our system outperforms existing methods in
           the high precision regime, and is virtually identical to SE at high
           recall.}
  \label{fig:pr}
\end{figure}

Figure~\ref{fig:pr} shows the performance of our model on the BSDS500 test set
over the full range of operating thresholds. Our system outperforms 
existing methods in the high precision regime, and is virtually identical to
SE~\cite{PiotrPAMI} at high recall. Table~\ref{tbl:bsds-full} lists
quantitative benchmark results and compares them to recently published methods.

\vspace{-0.3cm}
\paragraph{Regions}
We combine OEF with MCG~\cite{MCG} to produce segmentation hierarchies from our
edge detector output. MCG originally used contour strengths from SE, and we
found its implementation is sensitive to the statistics of SE output. Rather
than tune the implementation, we simply applied a monotonic transformation of
our detector output to match the SE distribution (see Section~\ref{sec:vis}).
The resulting combination, denoted OEF+MCG in Figure~\ref{fig:pr} and
Table~\ref{tbl:bsds-full}, is surprisingly effective, attaining an ODS of 0.76
on BSDS.

\vspace{-0.3cm}
\paragraph{Diagnostic experiments}
The performance benefits of calibration are shown in Table~\ref{tbl:variants}.
Calibration results in a clear improvement below 50\% recall, boosting average
precision from 0.81 to 0.82. In the same table we also report benchmark scores
for our model when predictions from the ensemble are combined by voting
(Eqn~\ref{eqn:vote}) rather than averaging. Voting appears to match averaging
up to roughly 20\% recall, beyond which it falls behind.

\vspace{-0.3cm}
\paragraph{Amount of training data}
We find that our model benefits significantly from large amounts of training
data. In Figure~\ref{fig:bigdata}, we show how performance on BSDS500 varies as
the amount of patches used for training is increased. Important for utilizing
large datasets is efficient training. We discuss timing details in
Section~\ref{sec:timing}.

\subsection{Visualizing Detector Output}
\label{sec:vis}

Qualitative results on a selection of test images are shown in
Figure~\ref{fig:montage}.  Notice that although the forest is trained only to
detect straight edges, its performance at corners and junctions is as good as
any other method.

\begin{figure}[h]
  \begin{center}
  \begin{tabular}{cc}
    \includegraphics[width=0.42\linewidth]{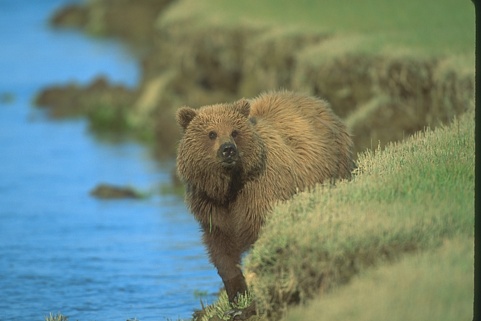} &
    \hspace{-0.4cm}
    \includegraphics[width=0.42\linewidth]{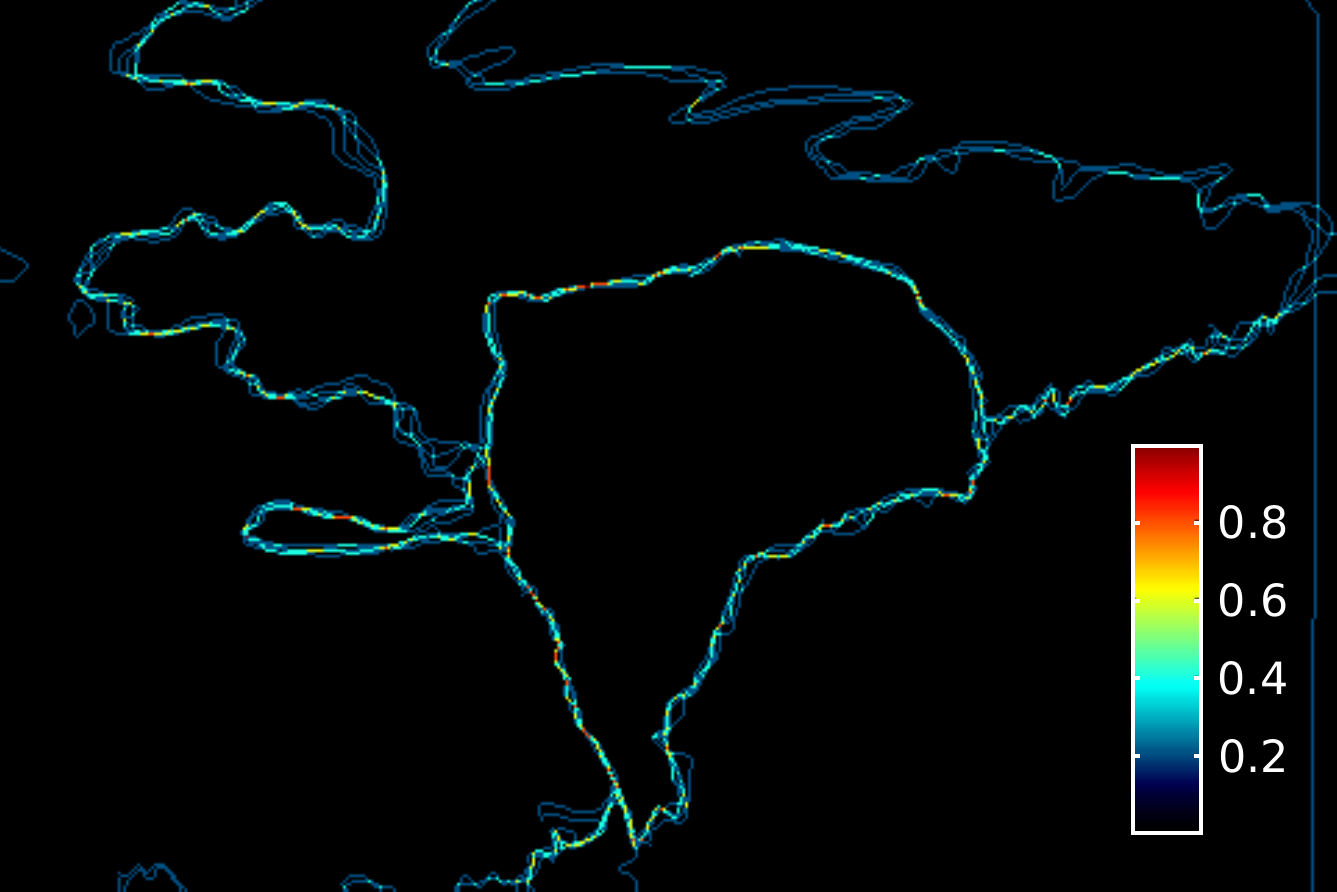} \\ \noalign{\vskip -0.8mm}
    (a) Original image & (b) Ground truth \\ \noalign{\vskip 1.3mm}
    \includegraphics[width=0.42\linewidth]{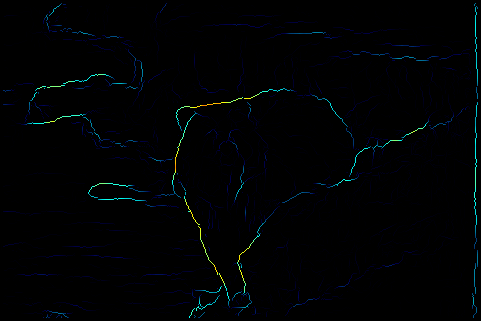} &
    \hspace{-0.4cm}
    \includegraphics[width=0.42\linewidth]{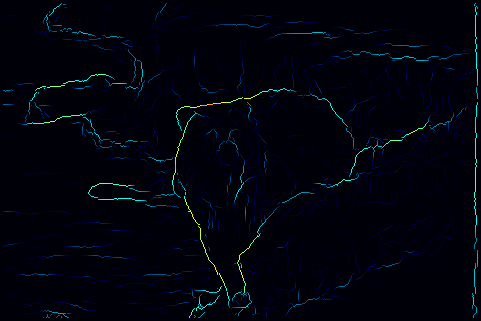} \\ \noalign{\vskip -0.8mm}
    (c) SCG original & (d) SCG new \\ \noalign{\vskip 1.3mm}
    \includegraphics[width=0.42\linewidth]{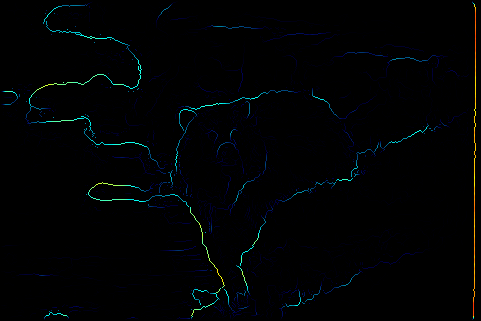} &
    \hspace{-0.4cm}
    \includegraphics[width=0.42\linewidth]{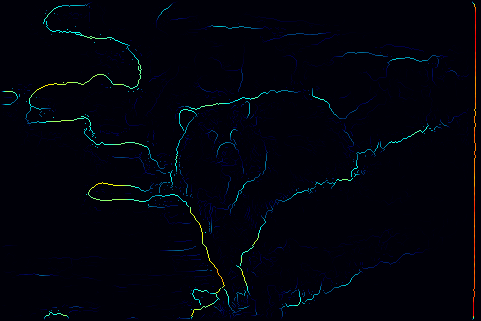} \\ \noalign{\vskip -0.8mm}
    (e) SE original & (f) SE new \\ \noalign{\vskip 1.3mm}
    \includegraphics[width=0.42\linewidth]{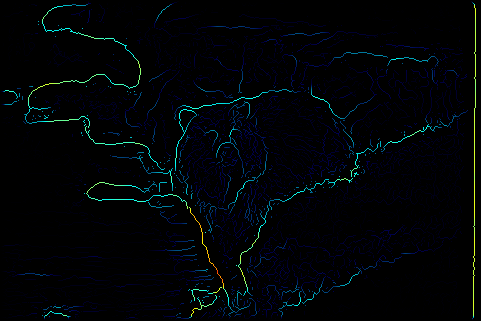} &
    \hspace{-0.4cm}
    \includegraphics[width=0.42\linewidth]{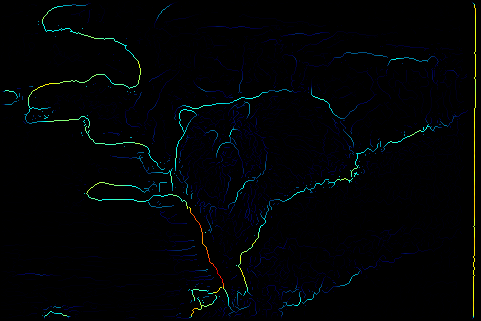} \\ \noalign{\vskip -0.8mm}
    (g) OEF original & (h) OEF new 
  \end{tabular}
  \end{center}
  \vspace{-0.2cm}
  \caption{%
  We compute a monotonic transformation for each algorithm which attempts to
  make the distributions of output values as similar as possible. This serves
  to highlight relevant differences by making the visualizations of different
  algorithms invariant to monotonic transformations that don't effect benchmark
  performance. Here we show the effect of normalizing all three algorithms to
  match the average SE distribution. Note that this even changes the appearance
  of SE since its distribution on a particular image will differ from its
  average distribution over all images.}
  \vspace{-0.25cm}
  \label{fig:histeq}
\end{figure}

\begin{figure}[t]
  \begin{center}
  \subfloat[ODS]{%
    \includegraphics[width=0.45\linewidth]{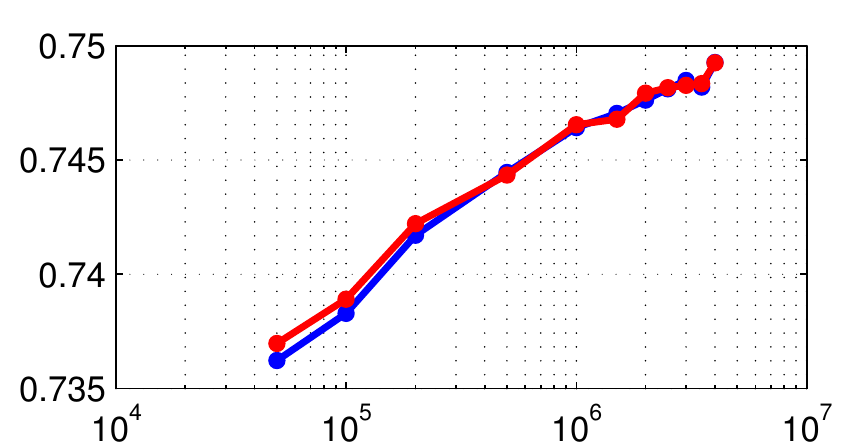}
  }%
  \hspace{.1cm}
  \subfloat[AP]{%
    \includegraphics[width=0.45\linewidth]{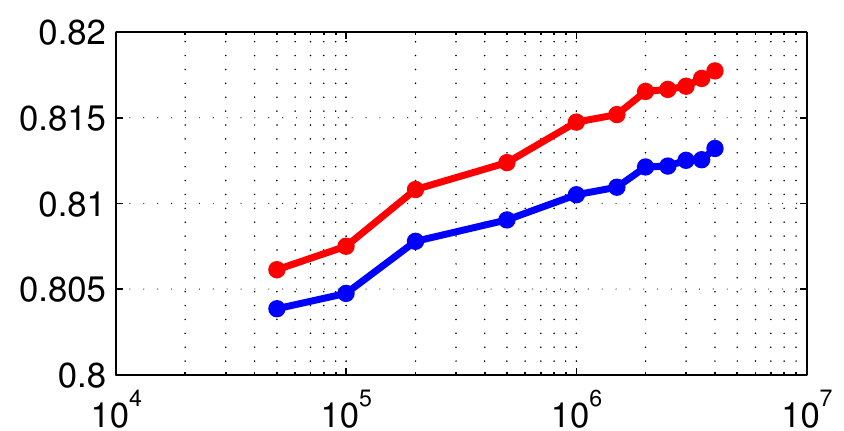}
  }%
  \end{center}
  \vspace{-0.1in}
  \caption{Performance on BSDS500 as a function of the number of training
           examples, before calibration ({\color{blue}blue}) and after
           calibration ({\color{red}red}). The smallest model was trained on
           $5\times10^4$ examples and the largest on $4\times10^6$ examples.
           Training times vary from less than one minute (40 seconds data
           collection $+$ 6 seconds tree training) per tree for the smallest
           model to under 30 minutes (15-20 minutes data collection $+$ 5
           minutes tree training) per tree for the largest model.}
\label{fig:bigdata}
\end{figure}

\begin{figure}
  \begin{center}
    \includegraphics[width=0.8\linewidth]{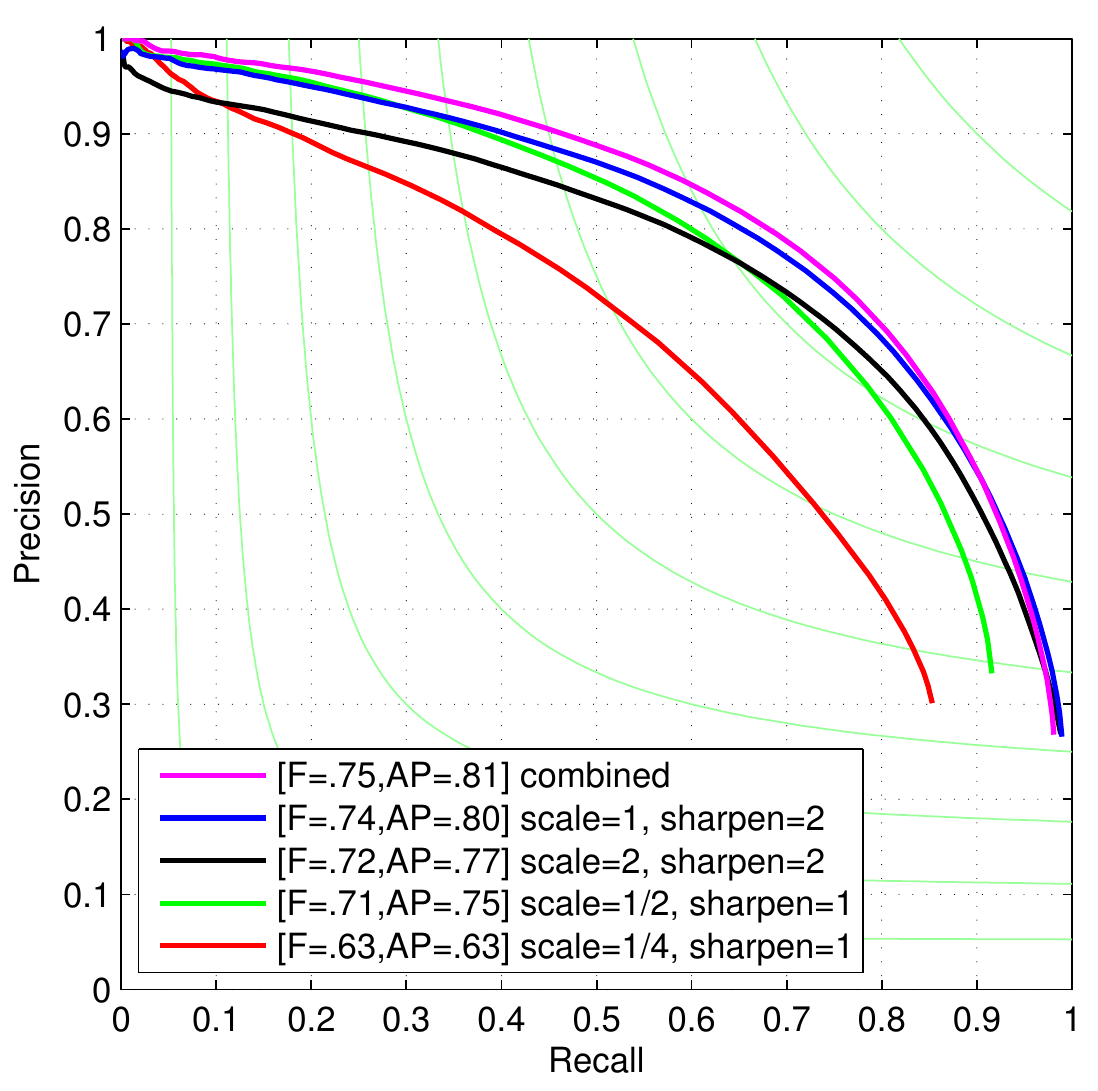}
    \vspace{-0.1in}
  \end{center}
  \caption{Results on BSDS showing the performance of our algorithm when run at
           a particular scale, compared to the results after multiscale
           combination. No calibration is performed here. Consistent with the
           findings of \cite{ren2008multi}, the combined model greatly
           outperforms any fixed-scale model.}
\label{fig:scalePR}
\end{figure}

\begin{figure*}[t]
  \captionsetup[subfigure]{labelformat=empty}
  \hspace{-0.6cm}
  \centering
  \subfloat[(a) Original image]{%
    {\renewcommand{\arraystretch}{0.7}
    \begin{tabular}{c}
      \includegraphics[width=0.195\textwidth]{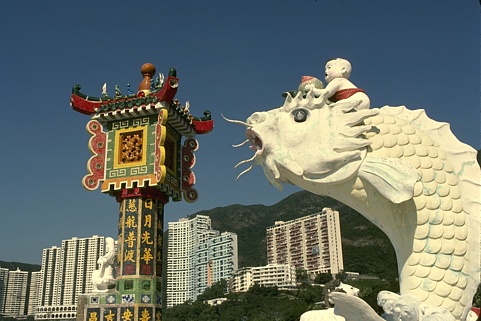} \\
      \includegraphics[width=0.195\textwidth]{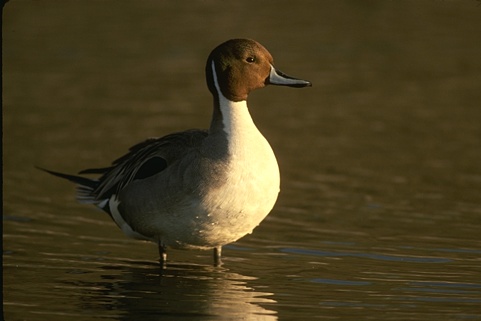} \\
      \includegraphics[width=0.195\textwidth]{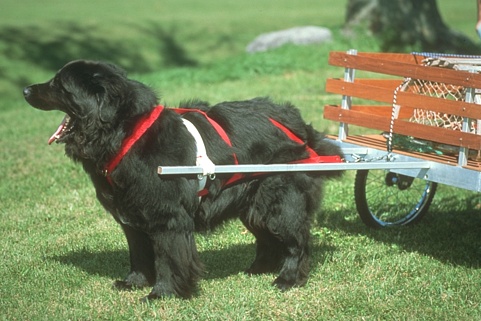} \\
      \includegraphics[width=0.195\textwidth]{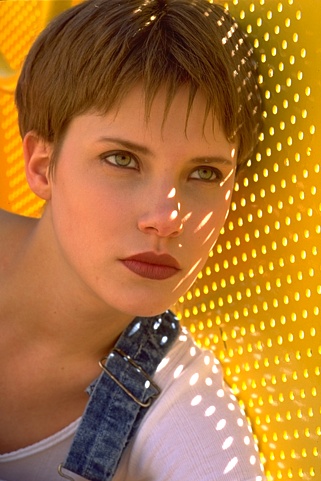} \\
      \includegraphics[width=0.195\textwidth]{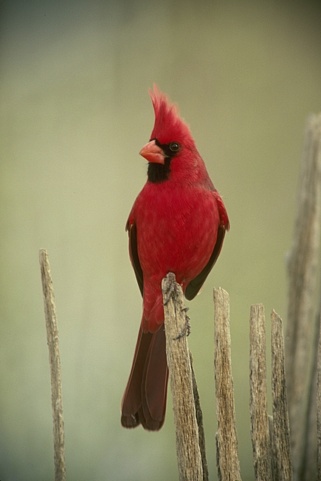}
    \end{tabular}} \hspace{-0.6cm}
  }%
  \subfloat[(b) Ground truth]{%
    {\renewcommand{\arraystretch}{0.7}
    \begin{tabular}{c}
      \includegraphics[width=0.195\textwidth]{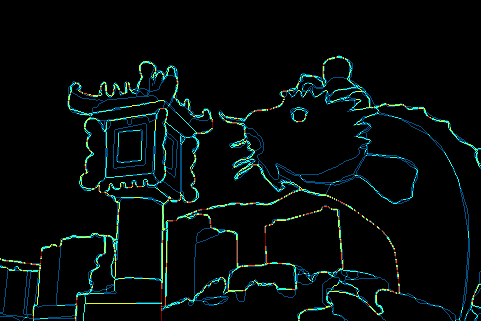} \\
      \includegraphics[width=0.195\textwidth]{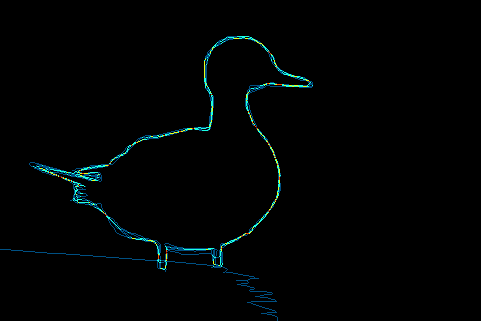} \\
      \includegraphics[width=0.195\textwidth]{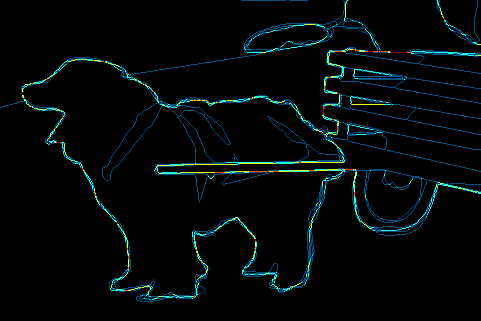} \\
      \includegraphics[width=0.195\textwidth]{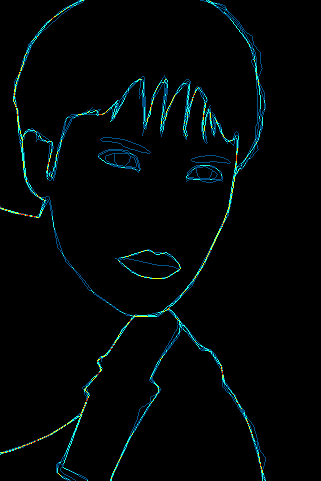} \\
      \includegraphics[width=0.195\textwidth]{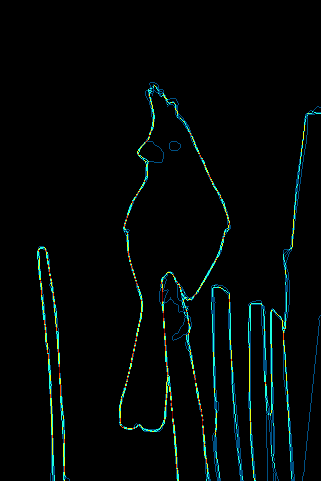}
    \end{tabular}} \hspace{-0.6cm}
  }%
  \subfloat[(c) SCG \cite{SCG}]{%
    {\renewcommand{\arraystretch}{0.7}
    \begin{tabular}{c}
      \includegraphics[width=0.195\textwidth]{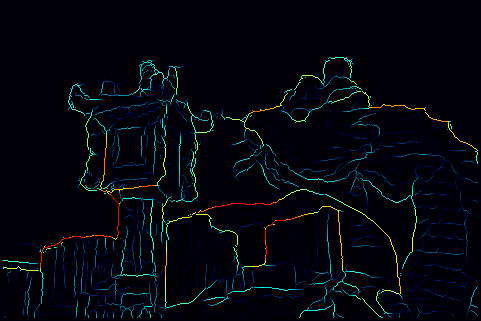} \\
      \includegraphics[width=0.195\textwidth]{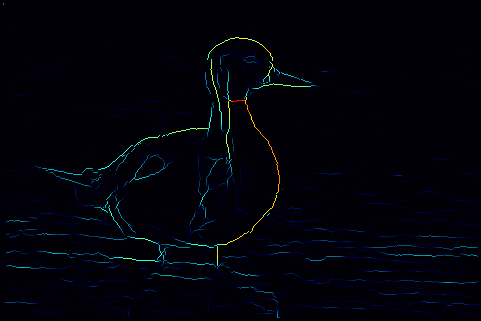} \\
      \includegraphics[width=0.195\textwidth]{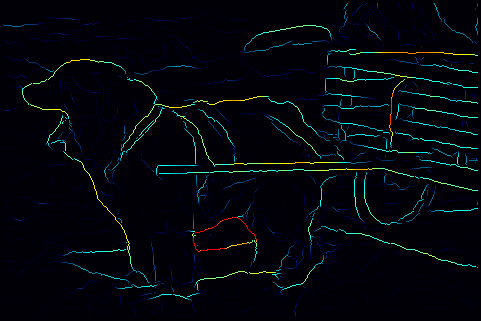} \\
      \includegraphics[width=0.195\textwidth]{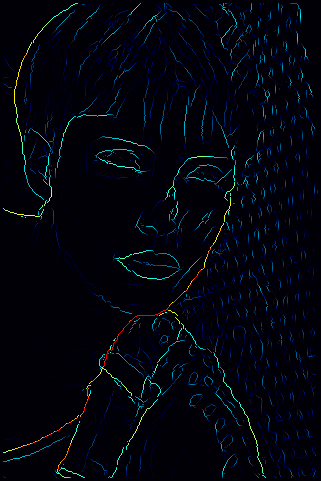} \\
      \includegraphics[width=0.195\textwidth]{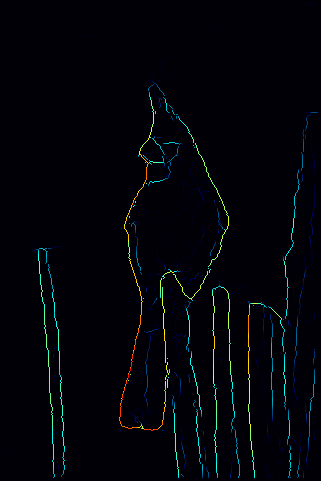}
    \end{tabular}} \hspace{-0.6cm}
  }%
  \subfloat[(d) SE \cite{PiotrPAMI}]{%
    {\renewcommand{\arraystretch}{0.7}
    \begin{tabular}{c}
      \includegraphics[width=0.195\textwidth]{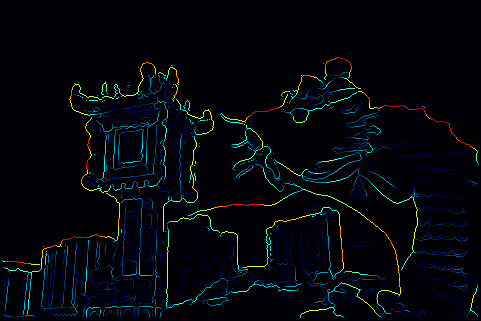} \\
      \includegraphics[width=0.195\textwidth]{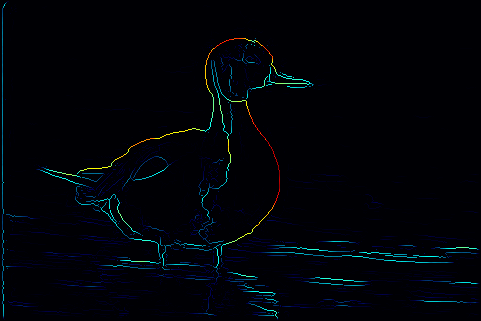} \\
      \includegraphics[width=0.195\textwidth]{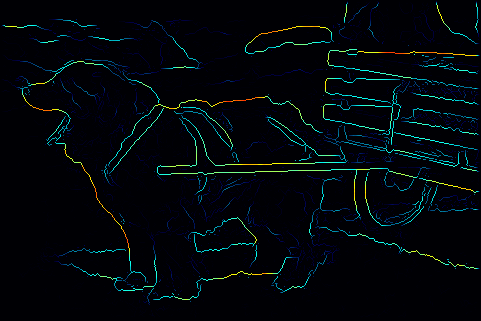} \\
      \includegraphics[width=0.195\textwidth]{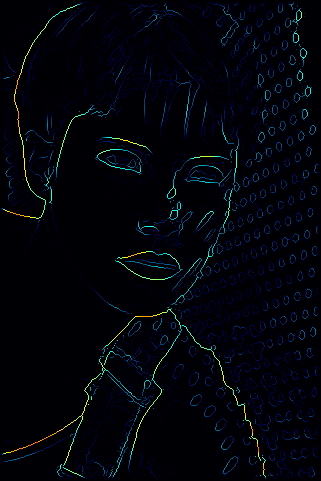} \\
      \includegraphics[width=0.195\textwidth]{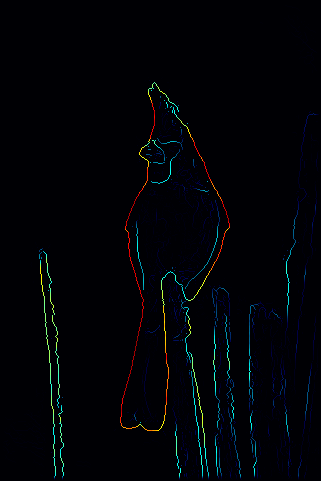}
    \end{tabular}} \hspace{-0.6cm}
  }%
  \subfloat[(e) OEF]{%
    {\renewcommand{\arraystretch}{0.7}
    \begin{tabular}{c}
      \includegraphics[width=0.195\textwidth]{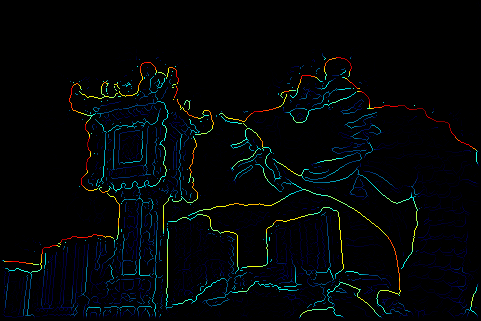} \\
      \includegraphics[width=0.195\textwidth]{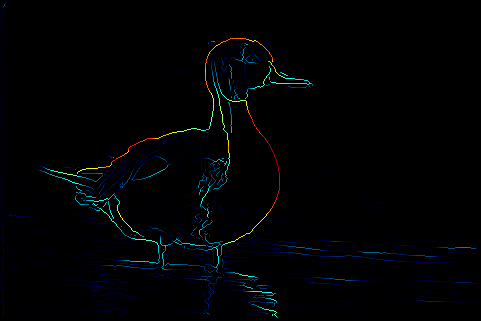} \\
      \includegraphics[width=0.195\textwidth]{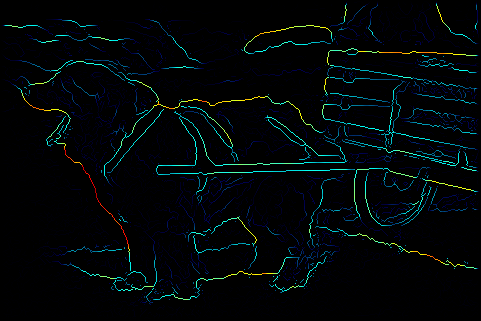} \\
      \includegraphics[width=0.195\textwidth]{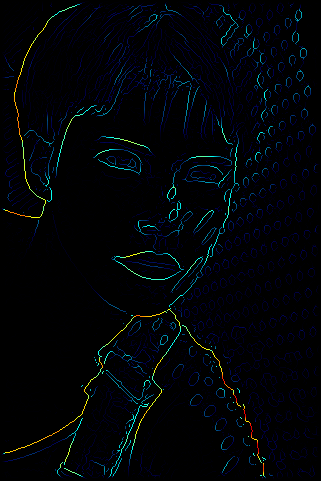} \\
      \includegraphics[width=0.195\textwidth]{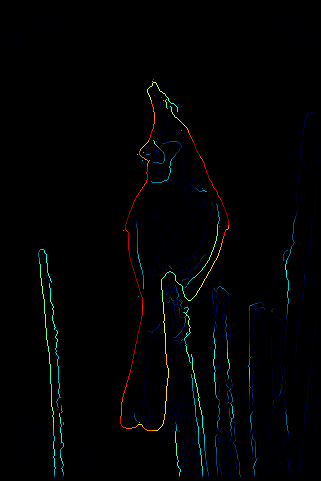} 
    \end{tabular}} \hspace{-0.6cm}
  }%
  \vspace{.4cm}
  \caption{Example results on the BSDS test set after non-maximal suppression.
           Rows 1,4 demonstrate our model correctly suppressing edges belonging
           to background texture, such as on the scales on the statue and the
           dots around the woman's face. Also note that in row 2 our results
           show significantly less weight on the false edges along the surface
           of the water. To allow for meaningful visual comparisons, we derive
           a global monotonic transformation for each algorithm that attempts to
           make the distributions of output values the same across all
           algorithms.  This post-processing step preserves the relative
           ordering of the edges, so benchmark results are unaffected but some
           irrelevant differences are eliminated from the boundary map
           visualization. Details can be found in Section~\ref{sec:vis}.}
  \label{fig:montage}
\end{figure*}

One difficulty with visualizing boundary detector outputs is that monotonic
transformations of the output boundary maps do not affect benchmark performance
but can dramatically affect the qualitative perception of boundary quality.  A
consequence of this is that qualitative comparisons of different algorithms can
be misleading, as the most salient differences tend not to be relevant to
actual performance.
An example of this can be seen in Figure~\ref{fig:histeq}(c), which shows the
raw output of SCG~\cite{SCG} on a test image. Comparing this against the
results of other algorithms, the output appears overly dim and gives the
impression that SCG is relatively insensitive to all but the most obvious
edges. However, a quick glimpse at its precision-recall curve reveals that this
impression is false.

To visualize boundary detector outputs in a way that highlights relevant
differences but removes these nuisance factors without affecting benchmark
results, we determine a global monotonic transformation for each boundary
detector which attempts to make the average histogram of response values across
all images match a standard distribution.  We first choose a reference
algorithm (we used SE) and compute its histogram of responses over an image set
to arrive at a target distribution.  For every boundary map produced by another
algorithm we compute a monotonic transformation for that boundary map that
approximately matches its histogram to the target distribution. Averaging these
mappings produces a single monotonic transformation specific to that algorithm
which we use when displaying outputs.



\begin{table}
  \begin{center}
  \small
  \begin{tabular}{l|ccc}
     & ODS & OIS & AP\footnotemark
     \\
    \hline
    Human         & .80 & .80 &     \\
    \hline
    gPb \cite{gPbPAMI}                & .71 & .74 & .65 \\
    gPb-owt-ucm \cite{gPbPAMI}        & .73 & .76 & .73 \\
    Sketch Tokens \cite{SketchTokens} & .73 & .75 & .78 \\
    SCG \cite{SCG}                    & .74 & .76 & .77 \\
    DeepNet \cite{deepnet}            & .74 & .76 & .76 \\
    PMI \cite{crisp}                  & .74 & .77 & .78 \\
    SE \cite{PiotrPAMI}               & .75 & .77 & .80 \\
    SE + MCG \cite{MCG}               & .75 & .78 & .76 \\
    \hline
    OEF        &         .75  &         .77  & \textbf{.82} \\
    OEF + MCG  & \textbf{.76} & \textbf{.79} & .76          \\
    \hline
  \end{tabular}
  \end{center}
  \vspace{-0.1in}
  \caption{Benchmark scores on BSDS500.}
  \label{tbl:bsds-full}
\end{table}

\footnotetext{We note that the lower AP for MCG is
because the benchmark computes average precision over the interval $[0,1]$
but the precision-recall curve does not extend to 0 recall. Monotonically
extending the curve to the left (e.g., as is done in PASCAL VOC) yields
AP values of gPb-owt-ucm=0.76, SCG=0.78, SE+MCG=0.81, OEF+MCG=0.82}

\subsection{Computational Costs}
\label{sec:timing}

A key advantage of our simplified approach relative to SE~\cite{PiotrPAMI}
is significantly reduced resources required at training time. We report
training times for both systems assuming each tree is trained on its own
bootstrap sample of $4\times10^6$ patches.

\vspace{-0.3cm}
\paragraph{Training}
For both models, the data sampling stage takes \mytilde20 minutes per tree.
Because we expose the trees to smaller random feature sets, this takes
approximately 15 gigabytes (GB) of memory, compared to 33 GB for SE. To train
on this much data, SE takes over 3.25 hours per tree and requires about 54 GB
of memory. This is due to the per-node discretization step, where at every tree
node PCA is applied to descriptors derived from the training examples at that
node. In contrast, our approach is almost 40$\times$ faster, taking about 5
minutes per tree, with memory usage at roughly 19 GB.

\vspace{-0.3cm}
\paragraph{Detection}
%
%
%
We report runtimes for images of size $480\times320$ on an 8-core Intel i7-950.
A voting implementation of our system (Eqn~\ref{eqn:vote}) runs in about 0.7
seconds per image, compared to 0.4 seconds for SE. Runtime increases to 2
seconds when using averaging (Eqn~\ref{eqn:avg}).

The primary reason that averaging is slower is it requires more time for edge
sharpening since the predicted score vectors $\mathbf{w}$ are not sparse.  To
reduce the amount of computation spent on sharpening, we leverage the following
observation.  The same oriented edge will appear at different offsets $d$ across
neighboring windows.  The weights $\mathbf{w}$ for a given orientation can thus
all be aligned (e.g., with the $d=0$ channel) by simple translation and summed
prior to sharpening.  Thus the collection of 120-dimensional distributions
computed over the image are ``collapsed'' down to 8 dimensions, one per
orientation. This optimization reduces runtime from 11 seconds down to just 2
seconds, while dropping ODS and AP by less than 0.003.

\begin{table}
  \begin{center}
  \small
  \vspace{.1in} 
  \begin{tabular}{l|ccc}
     & ODS & OIS & AP \\
    \toprule
    vote        & .74 & .77 & .80 \\ 
    average     & .75 & .77 & .81 \\ \midrule
    vote+cal    & .75 & .77 & .81 \\
    average+cal & .75 & .77 & .82 \\
    + sharp$\,$=$\,$2,2,2,2 & .75 & .77 & .81 \\
    + sharp$\,$=$\,$1,1,1,1 & .75 & .77 & .81 \\
    + sharp$\,$=$\,$0,0,0,0 & .74 & .77 & .78 \\
    \bottomrule
  \end{tabular}
  \end{center}
  \caption{We analyze different variants of our system on BSDS. We use the
           notation ``sharp=a,b,c,d'' to indicate the sharpening levels used
           for scales 1/4, 1/2, 1, 2, respectively. All algorithms use
           sharpen=1,1,2,2 unless otherwise stated. Rows 1-2 compare voting
           (Eqn~\ref{eqn:vote}) and averaging (Eqn~\ref{eqn:avg}) prior to
           calibration, showing that having trees emit full distributions over
           labels is more powerful than casting single votes. Rows 3-4 show
           that calibration improves performance. The last four rows correspond
           to the calibrated model with different sharpening levels, and show
           that it helps to do less sharpening at lower scales.}
  \label{tbl:variants}
\end{table}

\section{Discussion}
\label{sec:discuss}

In many ways our oriented edge forest is similar to SCG in that we train a
classifier which predicts the boundary contrast at each hypothesized edge
orientation.  A chief difference is the addition of the $d$ parameter which
allows the classifier to make useful predictions even when it is not centered
directly over an edge.  For a traditional detector, points near a boundary also
tend to have high contrast but it is unclear whether they should constitute
positive or negative examples, and such training data is often discarded.

Our proposed system is also quite similar to SE and Sketch Tokens (it uses the
same features, choice of classifier, etc.). We find it interesting that the
inclusion of other types of output, such as junctions or parallel edges, is not
necessary. Such events are quite rare, so there is probably not enough training
data to really learn the appearance of more complicated local segmentations. In
fact we found that training SE without complex patches ($>$2 segments) worked
just as well.

A final observation is that having the classifier output patches may not be
necessary.  It is certainly computationally advantageous since a given pixel
receives votes from many more trees, but given enough trees, we find that
Sketch Tokens performs essentially as well when only predicting the probability
at the center pixel.  This suggests that the real value of structured outputs
for edge detection is in partitioning the training data in a way that
simplifies the task of the decision tree: breaking patches into different
clusters allows the tree to learn the appearance of each cluster separately
rather than having to discover the structure by mining through large quantities
of data.  We hypothesize that other types of supervisory information---\eg
curvature, depth of a surface from the camera, change in depth across an edge,
figure-ground orientation of a contour, material or object category of a
surface---may further simplify the job of the forest, allowing it to better fit
the data more readily than simply training on a larger set of undistinguished
patches.

\begin{figure*}[t]
  \captionsetup[subfigure]{labelformat=empty}
  \hspace{-0.6cm}
  \centering
  \subfloat[(a) Original image]{%
    {\renewcommand{\arraystretch}{0.7}
    \begin{tabular}{c}
      \includegraphics[width=0.195\textwidth]{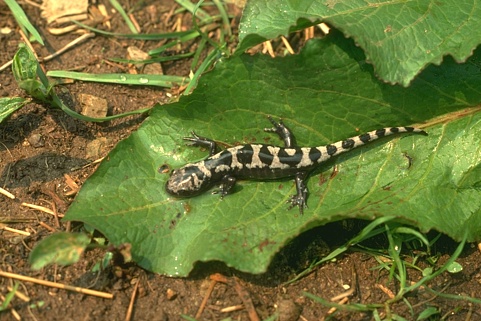} \\
      \includegraphics[width=0.195\textwidth]{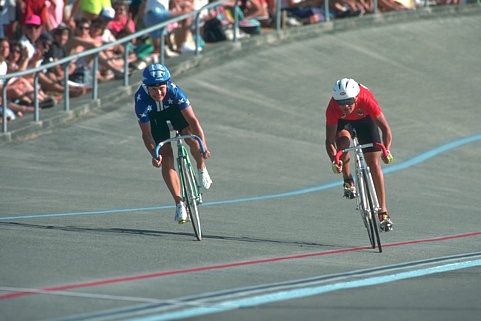} \\
      \includegraphics[width=0.195\textwidth]{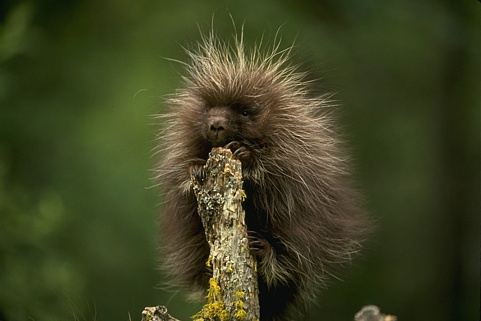} \\
      \includegraphics[width=0.195\textwidth]{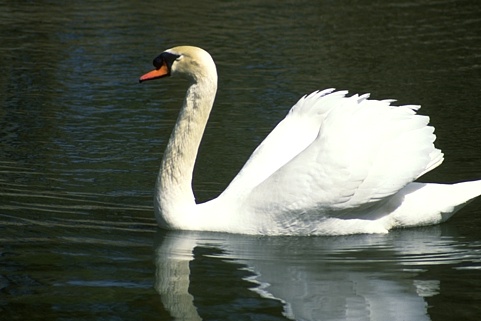} \\
      \includegraphics[width=0.195\textwidth]{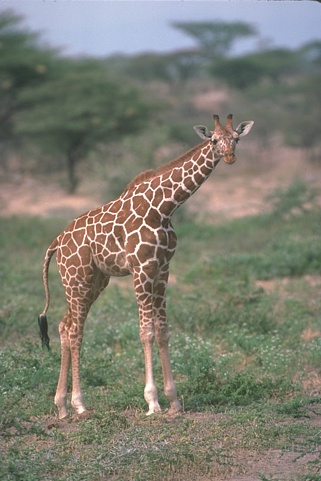} \\
      \includegraphics[width=0.195\textwidth]{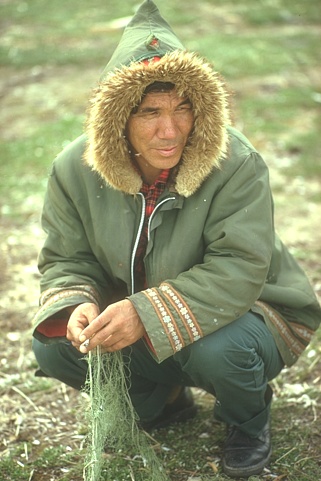} \\
    \end{tabular}} \hspace{-0.6cm}
  }%
  \subfloat[(b) Ground truth]{%
    {\renewcommand{\arraystretch}{0.7}
    \begin{tabular}{c}
      \includegraphics[width=0.195\textwidth]{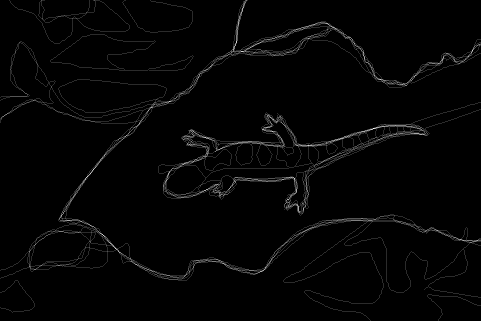} \\
      \includegraphics[width=0.195\textwidth]{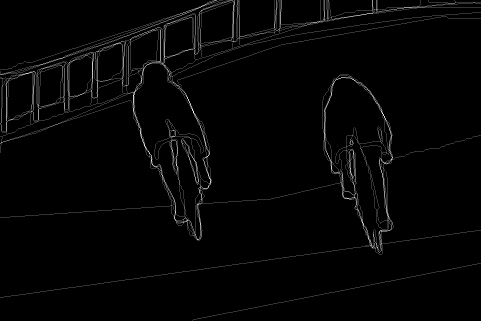} \\
      \includegraphics[width=0.195\textwidth]{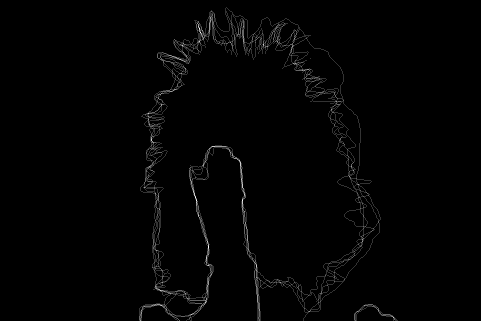} \\
      \includegraphics[width=0.195\textwidth]{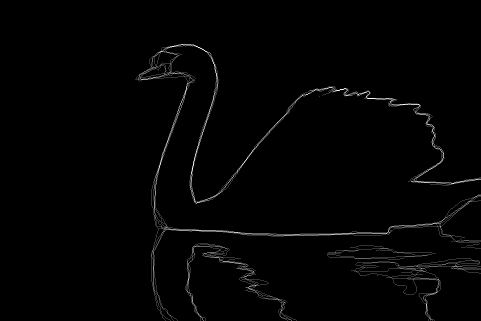} \\
      \includegraphics[width=0.195\textwidth]{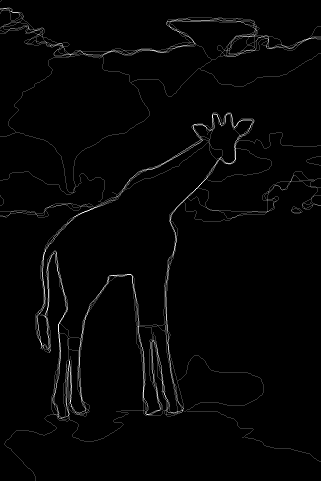} \\
      \includegraphics[width=0.195\textwidth]{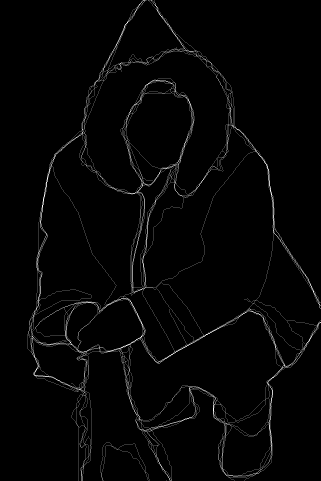} \\
    \end{tabular}} \hspace{-0.6cm}
  }%
  \subfloat[(c) gPb-owt-ucm \cite{gPbPAMI}]{%
    {\renewcommand{\arraystretch}{0.7}
    \begin{tabular}{c}
      \includegraphics[width=0.195\textwidth]{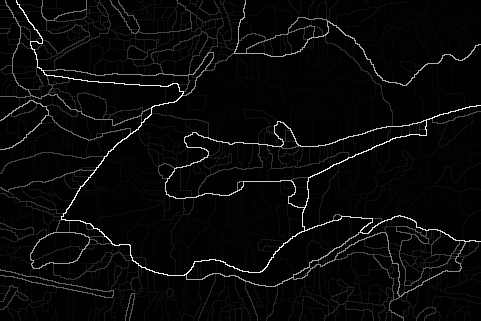} \\
      \includegraphics[width=0.195\textwidth]{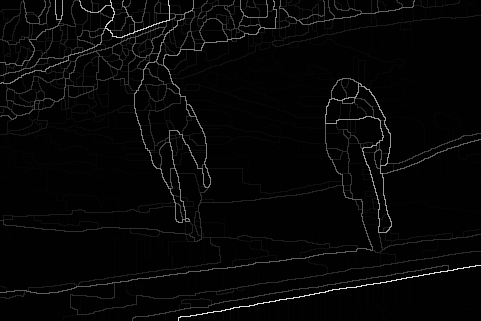} \\
      \includegraphics[width=0.195\textwidth]{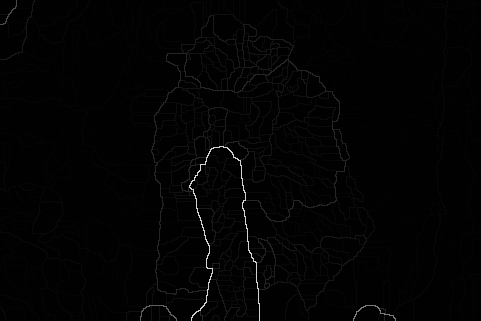} \\
      \includegraphics[width=0.195\textwidth]{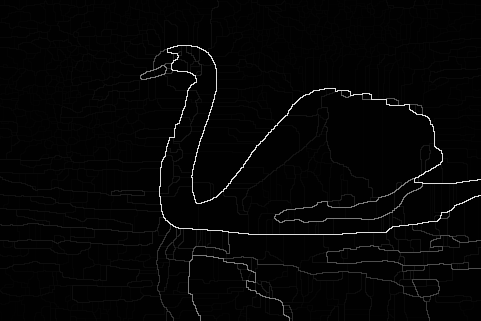} \\
      \includegraphics[width=0.195\textwidth]{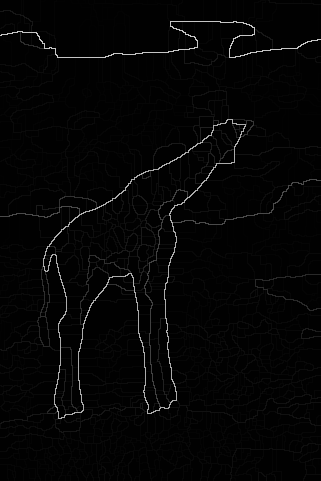} \\
      \includegraphics[width=0.195\textwidth]{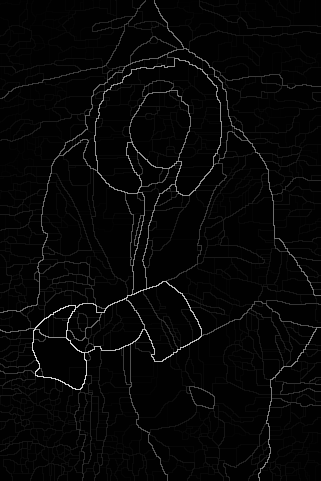} \\
    \end{tabular}} \hspace{-0.6cm}
  }%
  \subfloat[\qquad (d) SE \cite{PiotrJournal} + MCG \cite{MCG}]{%
    {\renewcommand{\arraystretch}{0.7}
    \begin{tabular}{c}
      \includegraphics[width=0.195\textwidth]{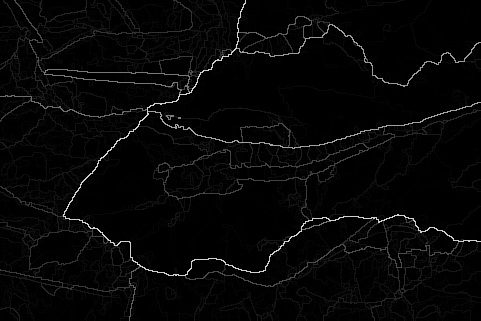} \\
      \includegraphics[width=0.195\textwidth]{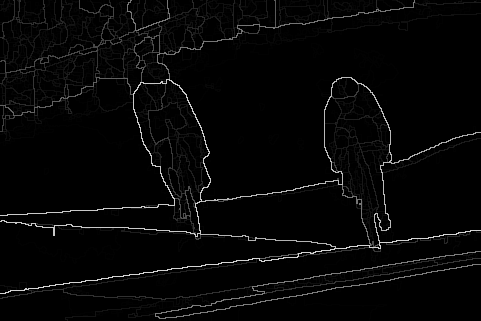} \\
      \includegraphics[width=0.195\textwidth]{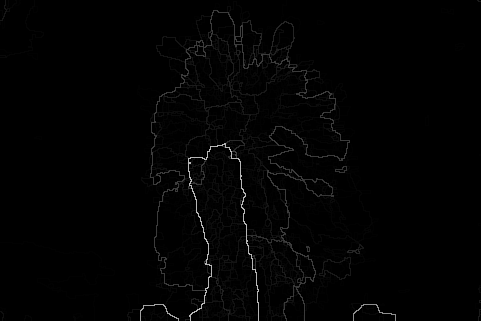} \\
      \includegraphics[width=0.195\textwidth]{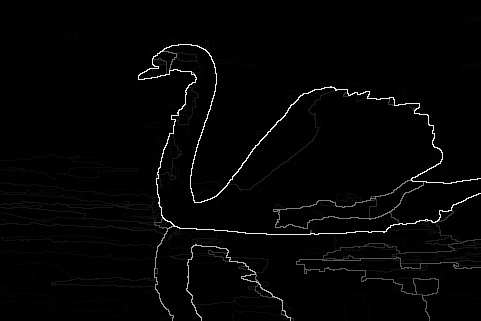} \\
      \includegraphics[width=0.195\textwidth]{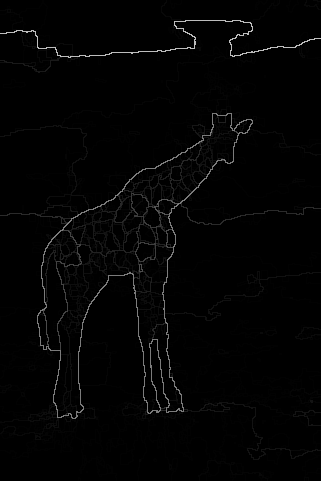} \\
      \includegraphics[width=0.195\textwidth]{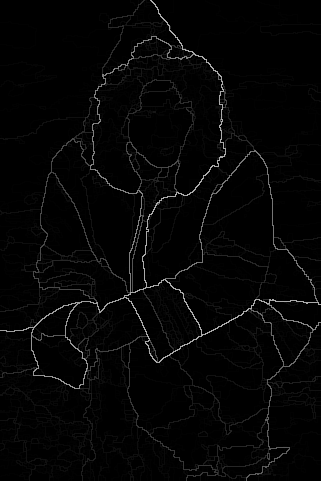} \\
    \end{tabular}} \hspace{-0.6cm}
  }%
  \subfloat[\qquad (e) OEF + MCG \cite{MCG}]{%
    {\renewcommand{\arraystretch}{0.7}
    \begin{tabular}{c}
      \includegraphics[width=0.195\textwidth]{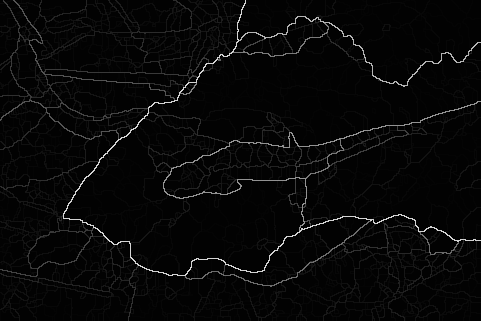} \\
      \includegraphics[width=0.195\textwidth]{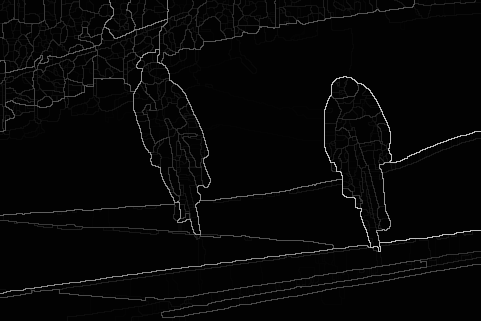} \\
      \includegraphics[width=0.195\textwidth]{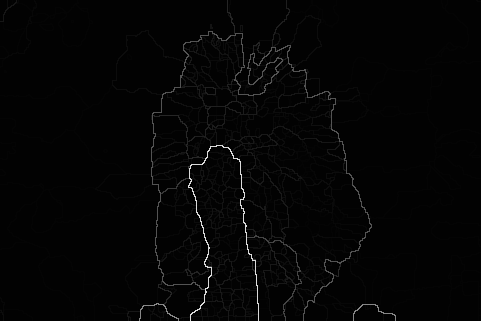} \\
      \includegraphics[width=0.195\textwidth]{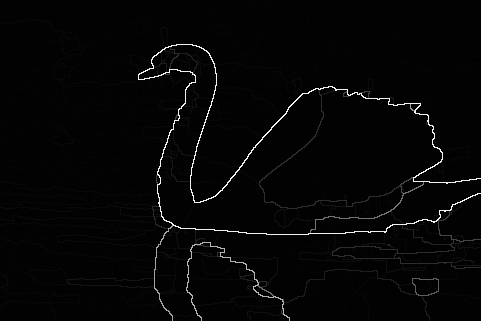} \\
      \includegraphics[width=0.195\textwidth]{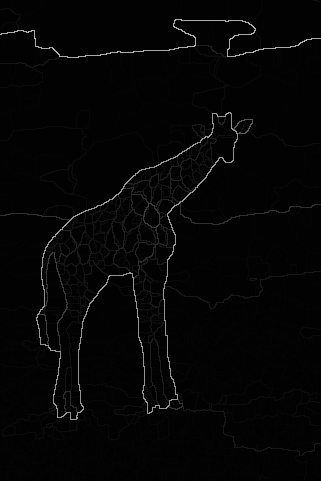} \\
      \includegraphics[width=0.195\textwidth]{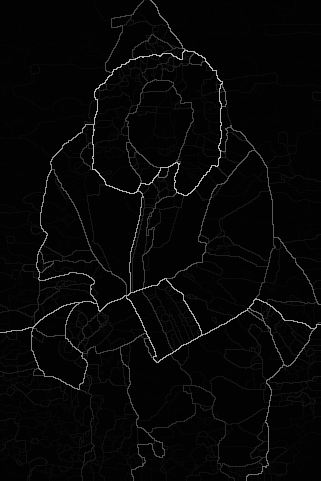} \\
    \end{tabular}} \hspace{-0.6cm}
  }%
  \vspace{.4cm}
  \caption{Segmentation hierarchies for a selection of BSDS test images,
  represented here as ultrametric contour maps \cite{gPbPAMI}. We found the
  combination of OEF and MCG~\cite{MCG} to produce high-quality segmentations
  that benchmark well on BSDS, achieving an F-measure of $0.76$. Details of
  this pipeline are given in Section~\ref{sec:bench} (\emph{Regions}). As in
  Figure~\ref{fig:montage}, results are visualized according to the procedure
  described in Section~\ref{sec:vis}.}
  \label{fig:segmontage}
\end{figure*}

{\footnotesize
\bibliographystyle{ieee}
\bibliography{refs}
}

\end{document}